**Dataset Descriptor**

PulseBat: A field-accessible dataset for second-life battery diagnostics from realistic histories using multidimensional rapid pulse test

**Author List**

Shengyu Tao[1], Guangyuan Ma[1], Huixiong Yang[2], Minyan Lu[2], Guodan Wei[1], Guangmin Zhou[1, *], Xuan Zhang[1, *]

**Author Affiliations**

1. Tsinghua Shenzhen International Graduate School, Tsinghua University, Shenzhen, China
2. Xiamen Lijing New Energy Technology Co., Ltd., Xiamen, China

**Significance Statement**

As electric vehicles (EVs) approach the end of their operational life, their batteries retain significant economic value and present promising opportunities for second-life use and material recycling. This is particularly compelling for Global South and other underdeveloped regions, where reliable energy storage is vital to addressing critical challenges posed by weak and even nonexistent power grid and energy infrastructures. However, despite this potential, widespread adoption has been hindered by critical uncertainties surrounding the technical performance, safety, and recertification of second-life batteries. In cases where they have been redeployed, mismatches between estimated and actual performance often render batteries technically unsuitable or hazardous, turning them into liabilities for communities they were intended to benefit. This considerable misalignment exacerbates energy access disparities and undermines the broader vision of energy justice, highlighting an urgent need for robust and scalable solutions to unlock the potential. In the PulseBat Dataset, the authors tested 464 retired lithium-ion batteries, covering 3 cathode material types, 6 historical usages, 3 physical formats, and 6 capacity designs. The pulse test experiments were performed repeatedly for each second-life battery with 10 pulse width, 10 pulse magnitude, multiple state-of-charge, and state-of-health conditions, e.g., from 0.37 to 1.03 (larger than the nominal capacity due to manufacturing inconsistencies). The PulseBat Dataset recorded these test conditions and the voltage response as well as the temperature signals that were subject to the injected pulse current, which could be used as a valuable data resource for critical diagnostics tasks such as state-of-charge estimation, state-of-health estimation, cathode material type identification, open-circuit voltage reconstruction, thermal management, and beyond. Part of the PulseBat Dataset was used in Nature Communications publications that addressed the state-of-health estimation problem under randomly distributed state-of-charge conditions[1].

**Correspondence**

Xuan Zhang: xuanzhang@sz.tsinghua.edu.cn; Guangmin Zhou: guangminzhou@sz.tsinghua.edu.cn

# 1 Overview

The overview of second-life batteries tested is summarized below:

| Batch | Material | Q (Ah) | Format | History | Quantity | State of health Range (Median, Std.) |
|---|---|---|---|---|---|---|
| 1 | NMC | 2.1 | Cylinder | Accelerated aging | 67 | 0.61-0.92 (0.83, 0.07) |
| 1 | LMO | 10.0 | Pouch | HEV1 | 95 | 0.51-0.95 (0.88, 0.11) |
| 1 | NMC | 21.0 | Pouch | BEV1 | 52 | 0.75-1.01 (0.99, 0.05) |
| 1 | LFP | 35.0 | Prismatic | HEV2 | 56 | 0.74-0.96 (0.85, 0.05) |
| 2 | LMO | 25.0 | Pouch | PHEV1 | 96 | 0.37-0.95 (0.80, 0.12) |
| 2 | LMO | 26.0 | Pouch | HEV3 | 98 | 0.78-1.03 (0.94, 0.07) |

where, NMC, LMO, and LFP stand for lithium nickel manganese cobalt oxide, lithium manganese oxide, and lithium iron phosphate, respectively. HEV, BEV, and PHEV stand for hybrid, battery, and plug-in hybrid electric vehicles, respectively. Q stands for the nominal capacity rated by the manufacturer. State-of-health value can be larger than 1 due to inconsistencies from the manufacturing process.

# 2 Experimental procedures

Tests were performed with BAT-NEEFLCT-05300-V010, NEBULA, Co, Ltd, and the air conditioner temperature was set at 25°C.

## 2.1 Step 1: Capacity Calibration

We use the widely adopted constant current (CC) discharge method as the gold standard for determining the capacity of retired batteries. Even considering the different initial state of charge (SOC) distributions of retired batteries, we use a unified method of first constant current constant voltage (CCCV) charging and then CC discharging to determine the capacity of retired batteries.

First, the retired batteries are charged to the upper cut-off voltage using a 1C constant current, then charged using constant voltage until the current drops to 0.05 C. The batteries are then discharged to the lower cut-off voltage using a 1C constant current. We use the actual discharge capacity as the calibrated (true) battery capacity and then let the battery rest for 20 minutes before SOC conditioning and pulse injection. The term C refers to the C-rate, determined by the current value required by a 1-hour full charge or discharge of a battery. The sampling frequency during Step 1 is 1 Hz. The cut-off conditions of Step 1 are listed in the table below:

| Batch | Material | Q (Ah) | Format | Cut-off voltage for discharging/charging (V) |
| --- | --- | --- | --- | --- |
| 1 | NMC | 2.1 | Cylinder | 2.0/4.2 |
| 1 | LMO | 10.0 | Pouch | 2.7/4.2 |
| 1 | NMC | 21.0 | Pouch | 2.7/4.2 |
| 1 | LFP | 35.0 | Prismatic | 2.5/3.65 |
| 2 | LMO | 25.0 | Pouch | 2.7/4.2 |
| 2 | LMO | 26.0 | Pouch | 2.7/4.2 |

## 2.2  Step 2: SOC Conditioning

SOC conditioning refers to adjusting the battery SOC to a desired level. The battery is at its zero SOC when the capacity calibration is finished. When a 5% SOC is desired, we use a 1C constant current for 3 minutes to adjust the calibrated battery to a 5% SOC level. The battery is then left to stand for 10 minutes to rest, expecting the battery to return to a steady state in preparation for subsequent pulse injection. Notice that SOC here is defined as the ratio of charged or dischargeable capacity to the nominal capacity. The sampling frequency during Step 2 is 1 Hz.

## 2.3  Step 3: Pulse Injection

The pulse width and pulse resting time are shown in the following table; that is, for each pulse width and resting time (each row of the table), we consecutively perform pulse injection with pulse amplitude being 0.5-1-1.5-2-2.5(C) in order, including *positive* and *negative* pulse injections. Note that positive and negative pulses alternate to cancel the equivalent energy injection; thus, the stored energy inside the battery does not change. Take pulse injection at 5% SOC as an example; at the 30ms pulse width, we inject 0.5C positive current pulse, then let the battery rest for 450ms, and then inject 0.5C negative current pulse, then again let the battery rest for 450ms. Still at 5% SOC, other remaining pulses with other amplitudes follow the rest time of the previous pulses. Repetitive experiments are performed until the remaining pulse widths are exhausted. Then, we charge the retired battery with a constant current of 1C for another 3 minutes to 10% SOC (refer to Step 1 for details), followed by the same procedure as explained above.

| Pulse width (ms) | Pulse rest time (ms) | Pulse magnitude (±C) |
|---|---|---|
| 30 | 450 | |
| 50 | 750 | |
| 70 | 1,050 | |
| 100 | 1,500 | |
| 300 | 4,500 | 0.5-1-1.5-2-2.5 |
| 500 | 7,500 | |
| 700 | 10,500 | |
| 1,000 | 15,000 | |
| 3,000 | 45,000 | |
| 5,000 | 75,000 | |

Repeat Step 2 and Step 3 until the SOC conditioning region is exhausted. The sampling frequency during Step 3 is 100 Hz.

### 2.4 SOC Conditioning Range Determination

The range of SOC conditioning is determined by a calibrated SOH of the retired battery. Specifically, the upper bound of the SOC conditioning region is lower than the calibrated minimal SOH value of the retired battery by 0.05. For instance, when the retired battery has a previously calibrated SOH between 0.50 and 0.55, then the SOC conditioning region will be 5% to 45%, with a grain of 5%. Detailed information is shown in the table below.

| State-of-health | State-of-charge (%), with a resolution of 5% |
|---|---|
| >0.95 | [5,90] |
| 0.90-0.95 | [5,85] |
| 0.85-0.90 | [5,80] |
| 0.80-0.85 | [5,75] |
| 0.75-0.80 | [5,70] |
| 0.70-0.75 | [5,65] |
| 0.65-0.70 | [5,60] |
| 0.60-0.65 | [5,55] |
| 0.55-0.60 | [5,50] |
| 0.50-0.55 | [5,45] |
| 0.45-0.50 | [5,40] |
| 0.40-0.45 | [5,35] |
| 0.35-0.40 | [5,30] |
| <0.35 | Not Found |

## 2.5 Voltage Protection

If the oscillation voltage during pulse injection exceeds the protection range, the current charging or discharging work step will be immediately terminated for a physical security check. If the security check is passed, no time will be made up for the already terminated work step, but the remaining work steps in the test procedure will be continued. In our test, voltage is mainly possible to exceed the protection range during charging, and no cases below the protection range during discharge have been found. The specific protection voltage parameters are consistent with those in the following table.

| Batch | Material | Q (Ah) | Format | Cut-off voltage for discharging/charging (V) |
|---|---|---|---|---|
| 1 | NMC | 2.1 | Cylinder | 1.95/4.3 |
| 1 | LMO | 10.0 | Pouch | 2.65/4.25 |
| 1 | NMC | 21.0 | Pouch | 2.65/4.3 |
| 1 | LFP | 35.0 | Prismatic | 2.45/3.7 |
| 2 | LMO | 25.0 | Pouch | 2.65/4.25 |
| 2 | LMO | 26.0 | Pouch | 2.65/4.25 |

## 2.6 SOC Deviation

The unequal charged and discharged capacity in adjacent positive and negative pulses with the same pulse intensity and planned pulse width caused by voltage protection will lead to an accumulative deviation in SOC to subsequent pulse tests. This SOC deviation is usually very slight due to the extremely short pulse width with no more than 5s. The voltage may exceed the protection range when the tested SOC is close to the SOH value of the battery. In Nature Communications publication[1], we only used data from 5-50% SOC. Considering that the SOH of the vast majority of batteries is above 0.6, the SOC deviation used can be ignored for simplicity. However, if readers want to use data at a higher SOC level, they need to pay attention to this SOC deviation issue to avoid introducing unnecessary errors.

# 3 Accessibility

The raw data and the manipulation code are accessible at this [link](link). The readers should cite the Nature Communications publication[1] and *this* data descriptor article when using the data.